\documentclass[10pt,twocolumn,letterpaper]{article}
\usepackage{pgfplots}
\usepackage{pgfkeys}
\usepackage{graphicx}
\usepackage{amssymb}
\usepackage{amsmath}
\usepackage{multirow}
\usepackage{picins}

\usepackage{xspace}
\newcommand*{\eg}{e.g.\@\xspace}
\newcommand*{\ie}{i.e.\@\xspace}
\newcommand{\bx}{\mathbf{x}}
\newcommand{\by}{\mathbf{y}}
\newcommand{\bz}{\mathbf{z}}
\newcommand{\bd}{\mathbf{d}}

\newcommand{\cnv}{\phi_{\text{cnv}}}
\newcommand{\fc}{\phi_{\text{fc}}}

\newcommand{\rcnn}{\text{RCNN}}
\newcommand{\gt}{\text{gt}}
\newcommand{\SPP}{\operatorname{SPP}}

\newcommand{\todo}[1]{}
\renewcommand{\todo}[1]{{\color{red} TODO: {#1}}}

\pgfplotsset{compat=newest}
\pgfplotsset{
	tick label style={font=\scriptsize},
	label style={font=\scriptsize},
	legend style={font=\scriptsize},
	title style={font=\scriptsize}}
\newlength\figureheight
\newlength\figurewidth

\renewcommand{\paragraph}[1]{\par\smallskip\noindent{\bf #1}}

\begin{document}
\title{R-CNN minus R}
\author{Karel Lenc ~~~~~~~~~~~~~~~~~~ Andrea Vedaldi\\
	\small Department of Engineering Science, University of Oxford\\
}
\maketitle
\begin{abstract}
Deep convolutional neural networks (CNNs) have had a major impact in most areas of image understanding, including object category detection. In object detection, methods such as R-CNN have obtained excellent results by integrating CNNs with region proposal generation algorithms such as selective search. In this paper, we investigate the role of proposal generation in CNN-based detectors in order to determine whether it is a necessary modelling component, carrying essential geometric information not contained in the CNN, or whether it is merely a way of accelerating detection. We do so by designing and evaluating a detector that uses a trivial region generation scheme, constant for each image. Combined with SPP, this results in an excellent and fast detector that does not require to process an image with algorithms other than the CNN itself. We also streamline and simplify the training of CNN-based detectors by integrating several learning steps in a single algorithm, as well as by proposing a number of improvements that accelerate detection.
\end{abstract}

\section{Introduction}\label{s:intro}

\begin{figure}[t]
\includegraphics[width=0.9\linewidth]{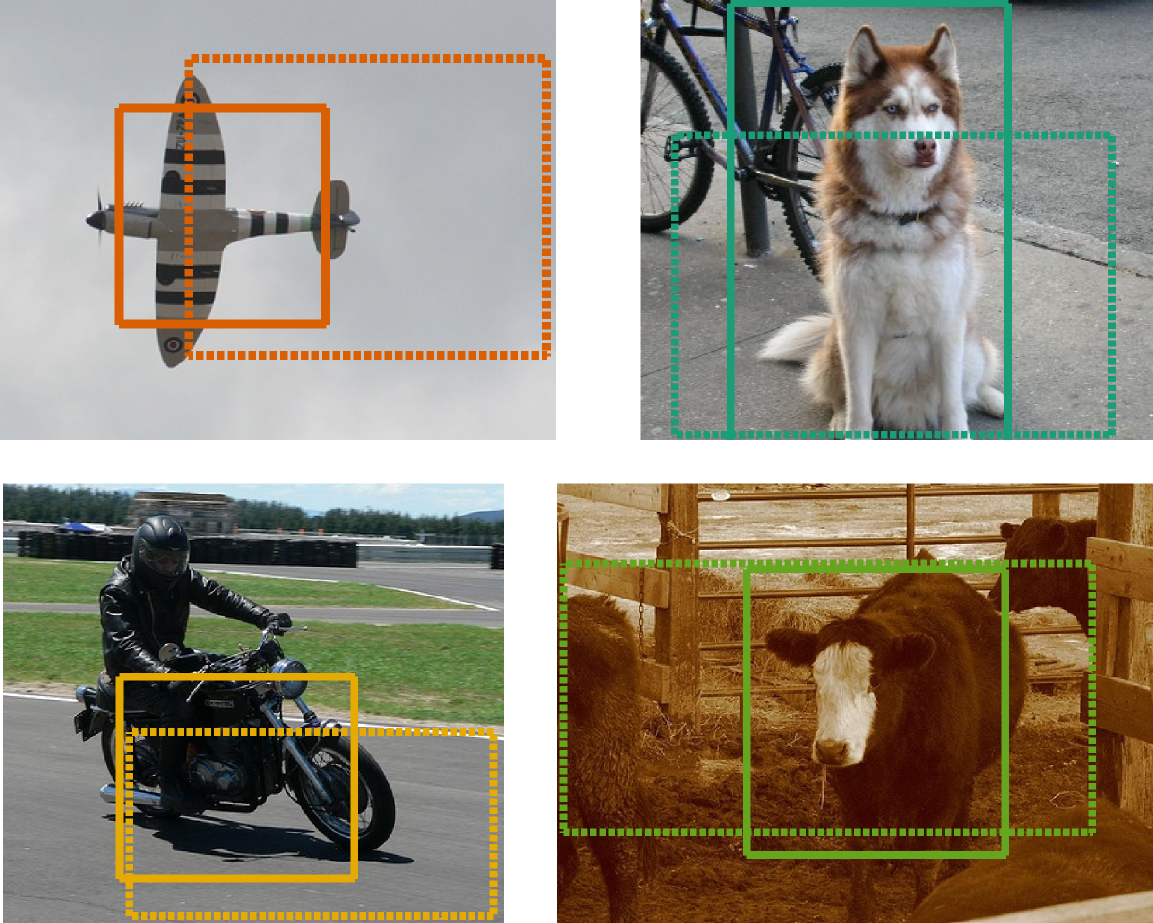}
\caption{Some examples of the bounding box regressor outputs. The dashed box is the image-agnostic proposal, correctly selected despite the bad overlap, and the solid box is the result of improving it by using the pose regressor. Both steps use the same CNN, but the first uses the geometrically-invariant fully-connected layers, and the last the geometry-sensitive convolutional layers. In this manner, accurate object location can be recovered \emph{without} using complementary mechanisms such as selective search.}\label{f:splash}
\end{figure}

Object detection is one of the core problems in image understanding. Until recently, the best performing detectors in standard benchmarks such as PASCAL VOC were based on a combination of handcrafted image representations such as SIFT, HOG, and the Fisher Vector and a form of structured output regression, from sliding window to deformable parts models. Recently, however, these pipelines have been outperformed significantly by the ones based on deep learning that acquire representations automatically from data using Convolutional Neural Networks (CNNs).  Currently,  the best CNN-based detectors are based on the R-CNN construction of~\cite{girshick14rich}. Conceptually, R-CNN is remarkably simple: it samples image regions using a proposal mechanism such as Selective Search (SS;~\cite{uijlings13selective}) and classifies them as foreground and background using a CNN. Looking more closely, however, R-CNN leaves open several interesting question.

The first question is whether CNN contain sufficient geometric information to localise objects, or whether the latter must be supplemented  by an external mechanism, such as region proposal generation. There are in fact two hypothesis. The first one is that the only role of proposal generation is to cut down computation by allowing to evaluate the CNN, which is expensive,  on a small number of image regions. In this case proposal generation becomes less important as other speedups such as SPP-CNN~\cite{gong14multi-scale} become available and may be forego. The second hypothesis, instead, is that proposal generation provides geometric information vital for accurate object localisation which is not represented in the CNN. This is not unlikely, given that CNNs are often trained to be highly invariant to even large geometric deformations and hence may not be sensitive to an object's location. This question is answered in Section~\ref{s:minusr} by showing that the convolutional layers of standard CNNs contain sufficient information to localise objects (Figure~\ref{f:splash}).

The second question is whether the R-CNN pipeline can be simplified. While conceptually straightforward, in fact, R-CNN comprises many practical steps that need to be carefully implemented and tuned to obtain a good performance. To start with, R-CNN builds on a CNN pre-trained on an image classification tasks such as ImageNet ILSVRC~\cite{deng09imagenet}.  This CNN is ported to detection by: i) learning an SVM classifier for each object class on top of the last fully-connected layer of the network, ii) fine-tuning the CNN on the task of discriminating objects and background, and iii) learning a bounding box regressor for each object class. Section~\ref{s:streamlined} simplifies these steps, which require running a mix of different software on cached data, by training a single CNN addressing all required tasks.

The third question is whether R-CNN can be accelerated. A substantial speedup was already obtained in \emph{spatial pyramid pooling} (SPP) by~\cite{he14spatial} by realising that convolutional features can be shared among different regions rather than being recomputed. However,  this does not accelerate training, and in testing the region proposal generation mechanism becomes the new bottleneck. The combination of dropping proposal generation and of the other simplifications are shown in Section~\ref{s:experiments} to provide a substantial detection speedup -- and this for the overall system, not just the CNN part. Our findings are summarised in Section~\ref{s:conclusions}.

\paragraph{Related work.} 
%
The basis of our work are the current generation of deep CNNs for image understanding, pioneered by~\cite{krizhevsky12imagenet}. For object detection, our method builds directly on the R-CNN approach of~\cite{girshick14rich} as well as the SPP extension proposed in~\cite{he04multiscale}. All such methods rely not only on CNNs, but also on a region proposal generation mechanism such as SS~\cite{uijlings13selective}, CPMC~\cite{carreira12cpmc}, multi-scale combinatorial grouping~\cite{arbelaez14multiscale},  and edge boxes~\cite{zitnick14edge}. These methods, which are extensively reviewed in~\cite{hosang15what}, originate in the idea of ``objectness'' proposed by~\cite{alexe10what}. Interestingly,~\cite{hosang15what} showed that a good region proposal scheme is essential for R-CNN to work well. Here, we show that this is in fact \emph{not} the case provided that bounding box locations are corrected by a strong CNN-based bounding box regressor, a step that was not evaluated for R-CNNs in~\cite{hosang15what}.
The R-CNN and SPP-CNN detectors build on years of research in object detection. Both can be seen as accelerated sliding window detectors~\cite{viola01rapid,dalal05histograms}. The two-stage computation using region proposal generation is a form of cascade detector~\cite{viola01rapid} or jumping window~\cite{sivic05discovering,vedaldi09multiple}. However, they differ in part-based detector such as~\cite{felzenszwalb08a-discriminatively} in that they do not explicitly model object parts in learning; instead parts are implicitly capture in the CNN.
Integrated training of SPP-CNN as a single CNN learning problem, not dissimilar to some of the ideas of Section~\ref{s:streamlined}, have very recently been explored in the unpublished manuscript~\cite{girshick15fast}.

\section{CNN-based detectors}\label{s:current}

This section introduces the R-CNN (Section~\ref{s:rcnn}) and SPP-CNN (Section~\ref{s:spp}) detectors.

\subsection{R-CNN detector}\label{s:rcnn}

The R-CNN method~\cite{girshick14rich} is a chain of conceptually simple steps: generating candidate object regions, classifying them as foreground or background,  and post-processing them to improve their fit to objects. These steps are described next.

\paragraph{Region proposal generation.} R-CNN starts by running an algorithm such as SS~\cite{uijlings13selective} or CPMC~\cite{carreira12cpmc} to extracts from an image $\bx$ a shortlist of of image regions $R \in \mathcal{R}(\bx)$ that are likely to contain objects. These proposals, in the order of a few thousands per image, may have arbitrary shapes, but in the following are assumed to be converted to rectangles.

\paragraph{CNN-based features.} Candidate regions are described by CNN features before being classified. The CNN itself is \emph{transferred} from a different problem -- usually image classification in the ImageNet ILSVRC challenge~\cite{deng09imagenet}. In this manner, the CNN can be trained on a very large dataset, as required to obtain good performance, and then applied to object detection, where datasets are usually much smaller.  In order to transfer a pre-trained CNN to object detection,  its last few layers, which are specific to the  classification task, are removed; this results in a ``beheaded'' CNN $\phi$ that outputs relatively generic features. The CNN is applied to the image regions $R$ by cropping and resizing the image $\bx$, \ie $\phi_{\text{RCNN}}(\bx;R) = \phi(\operatorname{resize} (\bx|_R))$. Cropping and resizing serves two purposes: to localise the descriptor and to provide the CNN with an image of a fixed size, as this is required by many CNN architectures.

\paragraph{SVM training.} Given the region descriptor $\phi_{\text{RCNN}}(\bx;R)$, the next step is to learn a SVM classifier to decide whether a region contains an object or background. Learning the SVM starts from a number of example images $\bx_1,\dots,\bx_N$, each annotated with ground-truth regions $\bar R \in \mathcal{R}_{\gt}(\bx_i)$ and object labels $c(\bar R) \in \{1,\dots C\}$. In order to learn a classifier for class $c$, R-CNN divides ground-truth $\mathcal{R}_{\gt}(\bx_i)$ and candidate $\mathcal{R}(\bx)$ regions into positive and negative. In particular, ground truth regions $R \in \mathcal{R}_{\gt}(\bx)$ for class $c(R) = c$ are assigned a positive label $y(R;c;\tau)=+1$; other regions $R$ are labelled as ambiguous $y(R;c;\tau)=\epsilon$ and ignored if $\operatorname{overlap}(R,\bar R)\geq \tau=0$ with any ground truth region $\bar R \in \mathcal{R}_{\gt}(\bx) $ of the same class $c(\bar R) = c$. The remaining regions are labelled as negative. Here $\operatorname{overlap}(A,B) = |A\cap B| / |A \cup B|$ is the intersection-over-union overlap measure, and the threshold is set to $\tau = 0.3$. The SVM takes the form $\phi_{\text{SVM}} \circ \phi_{\rcnn}(\bx;R)$, where $\phi_{\text{SVM}}$ is a linear predictor $\langle w_c, \phi_{\rcnn} \rangle + b_c$ learned using an SVM solver to minimise the regularised empirical hinge loss risk.

\paragraph{Bounding box regression.} Candidate bounding boxes are refitted to detected objects by using a CNN-based regressor as in~\cite{girshick14rich}. Given a candidate bounding box $R=(x,y,w,h)$, where $(x,y)$ are its centre and $(w,h)$ its width and height, a linear regressor estimates an adjustment $\bd = (d_x,d_y,d_w,d_h)$ that yields the new bounding box $\bd[R]= (w d_x + x, h d_y + y, w e^{d_w}, h e^{d_h})$. In order to train this regressor, one collects for each ground truth region $R^*$ all the candidates $R$ that overlap sufficiently with it (with an overlap of at least 0.5). Each pair $(R^*,R)$ of regions is converted in a training input/output pair $(\cnv(\bx,R),\bd)$ for the regressor, where $\bd$ is the adjustment required to transform $R$ into $R^*$, i.e. $R^* = \bd[R]$. The pairs are then used to train the regressor using ridge regression with a large regularisation constant. The regressor itself takes the form $\bd = Q_c^\top \cnv(\operatorname{resize}(\bx|_R)) + t_c$ where $\cnv$ denotes the CNN restricted to the convolutional layers, as further discussed in Section~\ref{s:spp}. The regressor is further improved by retraining it after removing the 20\% of the examples with worst regression loss -- as found in the publicly-available implementation of SPP-CNN.

\paragraph{Post-processing.} The refined bounding boxes are passed to non-maxima suppression before being evaluated. Non-maxima suppression eliminates duplicate detections prioritising regions with higher SVM score $\hat s(\phi(R))$. Starting from the highest ranked region in an image, other regions are iteratively removed if they overlap by more than $0.3$ with any region retained so far. 


\paragraph{CNN fine-tuning.} The quality of the CNN features, ported from an image classification task, can be improved by \emph{fine-tuning} the network on the target data. In order to do so, the CNN  $\phi_{\rcnn}(\bx;R)$ is concatenated with additional layers $\phi_{\text{sftmx}}$ (linear followed by softmax normalisation) to obtain a predictor for the $C+1$ object classes. The new CNN $\phi_{\text{sftmx}}\circ \phi_{\rcnn}(\bx;R)$ is then trained as a classifier by minimising its empirical logistic risk on a training set of labelled regions. This is analogous to the procedure used to learn the CNN in the first place, but with a reduced learning rate and a different (and smaller) training set similar to the one used to train the SVM. In this dataset, a region $R$, either ground-truth or candidate, is assigned the class $c(R;\tau_+,\tau_-) = c(\bar R^*)$ of the closest ground-truth region $\bar R^* = \operatornamewithlimits{argmax}_{\bar R \in \mathcal{R}_{\gt}(\bx)}\operatorname{overlap}(R, \bar R)$, provided that $\operatorname{overlap}(R,\bar R^*) \geq \tau_+$. If instead $\operatorname{overlap}(R,\bar R^*) < \tau_-$, then the region is labelled as $c(R;\tau_+,\tau_-) = 0$ background, and the remaining regions as ambiguous. By default $\tau_+$ and $\tau_-$ are both set $1/2$, resulting in a much more relaxed training set than for the SVM.  Since the dataset is strongly biased towards background regions, during CNN training it is rebalanced by sampling with 25\% probability regions such that $c(R) > 0$ and with 75\% probability regions such that $c(R)=0$.

\subsection{SPP-CNN detector}\label{s:spp}

A significant disadvantage of R-CNN is the need to recompute the whole CNN from scratch for each evaluated region; since this occurs thousands of times per image, the method is slow. SPP-CNN addresses this issue by factoring the CNN $\phi = \fc \circ \cnv$ in two parts, where $\cnv$ contains the so-called \emph{convolutional} layers, pooling information from local regions, and $\fc$ the \emph{fully connected} (FC) ones, pooling information from the image as a whole. 
Since the convolutional layers encode \emph{local information}, this can be selectively pooled to encode the appearance of an image subregion $R$ instead of the whole image. In more detail, let $\by = \cnv(\bx)$ the output of the convolutional layers applied to image $\bx$. The feature field $\by$ is a $H \times W \times D$ tensor of height $H$ and width $W$, proportional to the height and width of the input image $\bx$, and $D$ feature channels. Let $\bz = SP(\by;R)$ be the result of applying the \emph{spatial pooling} (SP) operator to the feature in $\by$ contained in region $R$. This operator is defined as:
\begin{equation}
	\label{e:sp}
  z_{d} = \max_{(i,j) : g(i,j) \in R} y_{ijd},\qquad d=1,\dots,D
\end{equation}
where the function $g$ maps the feature coordinates $(i,j)$ back to image coordinates $g(i,j)$. The SP operator is extended to \emph{spatial pyramid pooling} (SPP;~\cite{lazebnik06beyond}) by dividing the region $R$ into subregions $R = R_1 \cup R_2 \cup \dots R_K$, applying the SP operator to each, and then stacking the resulting features. In practice, SSP-CNN uses $K \times K$ subdivisions, where $K$ is chosen to match the size of the convolutional feature field in the original CNN. In this manner, the output can be concatenated with the existing FC layers:
$
\phi_{\SPP}(\bx;R)
=
\fc \circ \SPP(\cdot;R) \circ \cnv(\bx).
$
Note that, compared to R-CNN,  the first part of the computation is shared among all regions $R$. 

Next, we derive the map $g$ that transforms feature coordinates back to image coordinates as required by~\eqref{e:sp} (this correspondence was established only heuristically in~\cite{he14spatial}). It suffices to consider one spatial dimension. The question is which pixel $\bx_0(i_0)$ corresponds to feature $\bx_L(i_L)$ in the $L$-th layer of a CNN. While there is no unique definition, a useful one is to let $i_0$ be the \emph{centre of the receptive field} of feature $\bx_L(i_L)$, defined as the set of pixels $\Omega_L(i_L)$ that can affect $\bx_L(i_L)$ as a function of the image (i.e. the support of the feature). A short calculation leads to
\begin{align*}
i_0
& = 
g _L(i_L)
=
\alpha_L (i_L - 1) + \beta_L,
\\
\alpha_L & = \prod_{p=1}^L S_p,
\\
\beta_L
& = 
1
+ 
\sum_{p = 1}^L  \left(\prod_{q=1}^{p-1} S_q\right)
\left(
\frac{F_p - 1}{2}  - P_{p}
\right),
\end{align*}



\section{Simplifying and streamlining R-CNN}\label{s:method}


This section describes the main technical contributions of the paper: removing region proposal generation from R-CNN (Section~\ref{s:minusr}) and streamlining the pipeline (Section~\ref{s:streamlined}).

\subsection{Dropping region proposal generation}\label{s:minusr}

\begin{figure*}[t]
	\vspace{-3em}
	\begin{center}
		\begin{footnotesize}
			\setlength{\tabcolsep}{1pt}
			\newcommand{\CFImC}[1]{\includegraphics[height=0.15\linewidth]{figures/#1.pdf}}
			\begin{tabular}{ c c c c c}
			\hspace{3em}GT & SS & SS  $o_{GT} > 0.5$ & SW 7k & Cluster 3k  \\
     		\CFImC{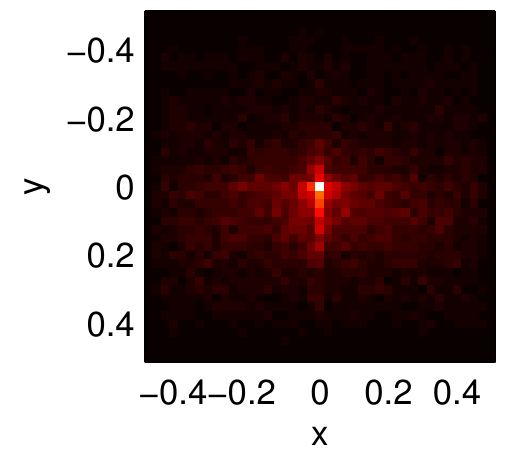} & \CFImC{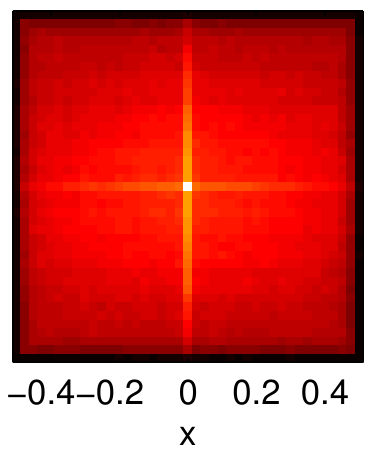} & \CFImC{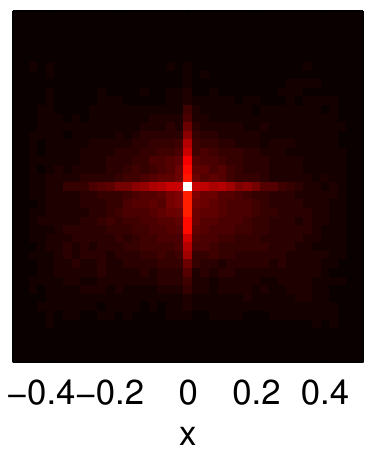} & \CFImC{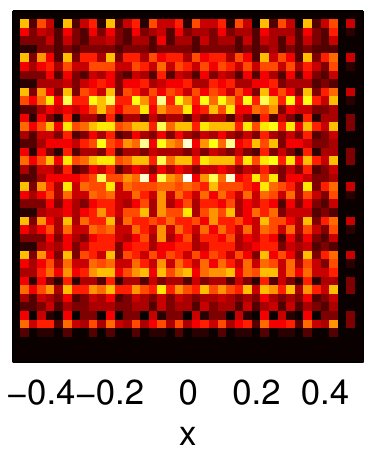} & \CFImC{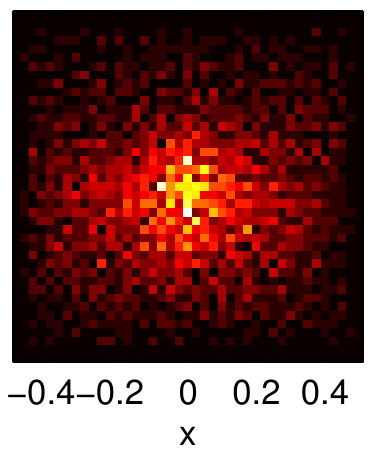}  \\
     		\CFImC{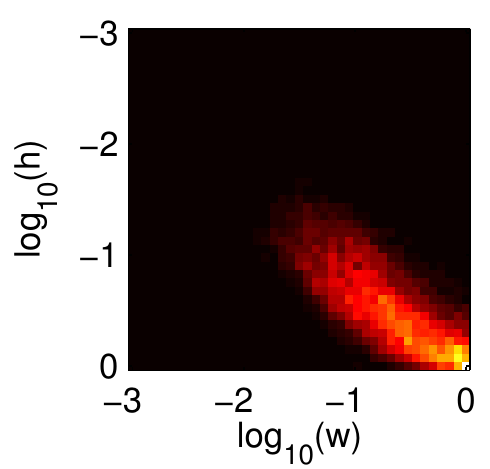} & \CFImC{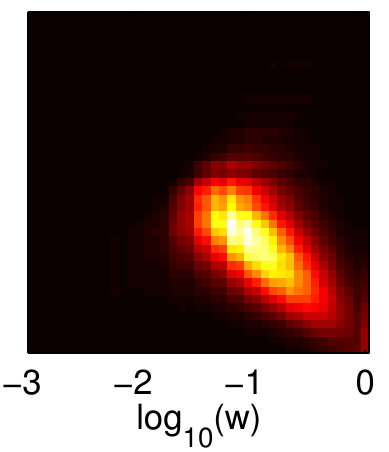} & \CFImC{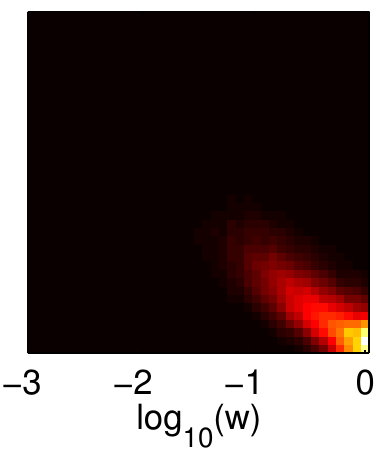} & \CFImC{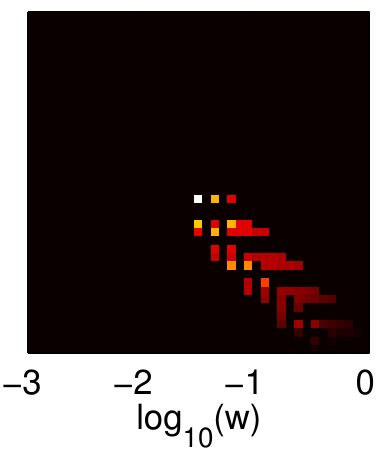} & \CFImC{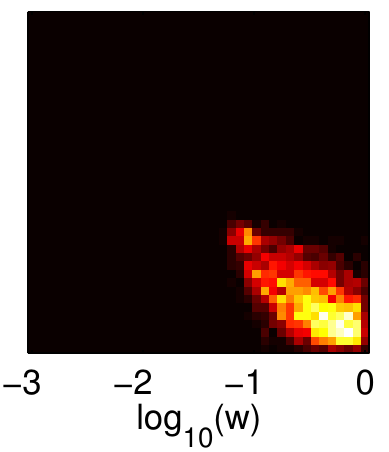}  \\
     		\CFImC{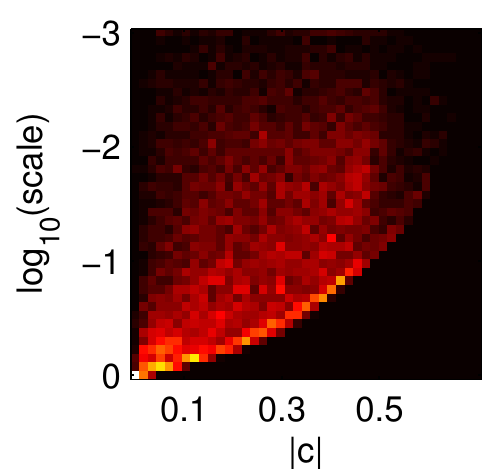} & \CFImC{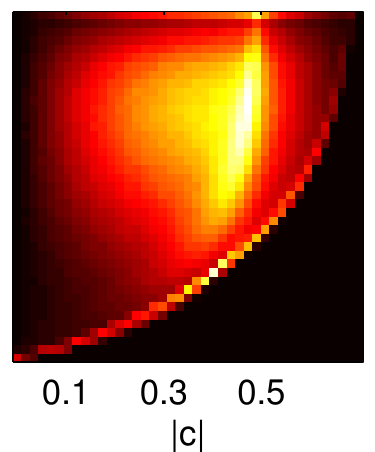} & \CFImC{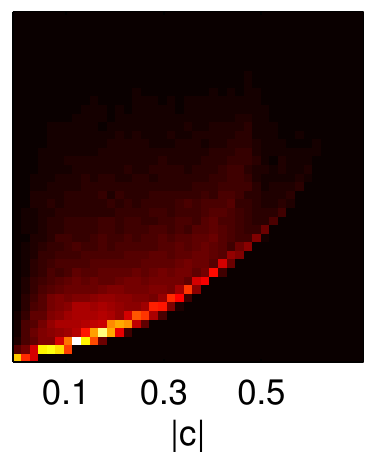} & \CFImC{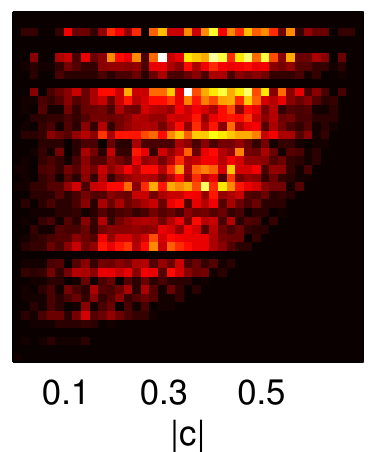} & \CFImC{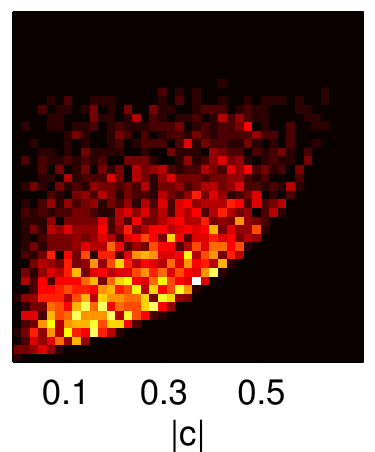}  \\
			\end{tabular}
			\caption{Bounding box distributions using the normalised coordinates of Section~\ref{s:minusr}. Rows show the histograms for the bounding box centre $(x,y)$, size $(w,h)$, and scale vs distance from centre $(s,|c|)$. Column shows the statistics for ground-truth, selective search, restricted selective search, sliding window, and cluster bounding boxes.}
			\label{fig:boxdist}
		\end{footnotesize}
	\end{center}
\end{figure*}

While the SPP method of~\cite{he14spatial} (Section~\ref{s:spp}) accelerates R-CNN evaluation by orders of magnitude, it does not result in a comparable acceleration of the detector as a whole; in fact, proposal generation with SS is about ten time slower than SPP classification. Much faster proposal generators exist, but may not result in very accurate regions~\cite{zhao14cracking}. Here we propose to drop $\mathcal{R}(\bx)$ entirely and to use instead an image-independent list of candidate regions $\mathcal{R}_0$, using the CNN itself to regress better object locations \emph{a-posteriori}.

Constructing $R_0$ starts by studying the distribution of bounding boxes in a representative object detection benchmark, namely the PASCAL VOC 207 data~\cite{everingham07pascal}.  A box is defined by the tuple $(r_s, c_s, r_e, c_e)$ denoting the upper-left and lower-right corners coordinates $(r_s,c_s)$ and $(r_e,c_e)$. Given an image of size $H \times W$, define the normalised with and height as $w = (c_e - c_e)/W$ and $h = (r_e - r_s)/H$ respectively; define also the scale $s = \sqrt{wh}$ and distance from the image centre $|c| = \| \left[(c_s + c_e)/2W - 0.5, (r_s + r_e)/2H - 0.5) \right] \|_2$.

 The first column of Figure~\ref{fig:boxdist} shows the distribution of such parameters for the GT boxes in the PASCAL data. It is evident that boxes tend to appear close to the image centre and to fill the image. The statistics of SS regions differs substantially; in particular, the $(s,|c|)$ histogram shows that SS boxes tend to distribute much more uniformly in scale and space compared to the GT ones. If SS boxes are restricted to the ones that have an overlap of at least 0.5 with a GT BB, then the distributions are similar again, with a strong preference for centred and large boxes.
 
The fourth column shows the distribution of boxes generated by a  \emph{sliding window} (SW;~\cite{dalal05histograms}) object detector. For an ``exhaustive'' enumeration of boxes at all location, scales, and aspect ratios, there can be hundred of thousands boxes per image. Here we subsample this set to 7K in order to obtain a candidate set with a size comparable to SS. This was obtained by sampling the width of the bounding boxes as $w = w_0 2^l, l =  0, 0.5, \dots 4$ where $w_0 \approx 40$ pixels is the width of the smallest bounding box considered in the SSP-CNN detector. Similarly, aspect ratios are sampled as $2^{\{ -1, -0.75, \dots 1 \}}$. The distribution of boxes, visualised in the fourth column of Figure~\ref{fig:boxdist}, is similar to SS and dissimilar from GT.


A simple modification of sliding window is to bias sampling to match the statistics of the GT bounding boxes. We do so by computing $n$ K-means clusters from the collection of vectors $(r_s, c_s, r_e, c_e)$ obtained from the GT boxes in the PASCAL VOC  training data. We call this set of boxes $\mathcal{R}_0(n)$; the fifth column of Figure~\ref{fig:boxdist} shows that, as expected, the corresponding distribution matches nicely the one of GT.  Section~\ref{s:experiments} shows empirically that, when combined with a CNN-based bounding box regressor, this proposal set results in a very competitive (and very fast) detector.
 	
\subsection{Streamlined detection pipeline}\label{s:streamlined}

This section proposes several simplifications to the R/SPP-CNN pipelines complementary to dropping region proposal generation as done in Section~\ref{s:minusr}. As a result of all these changes, the whole detector, including detection of multiple object classes and bounding box regression, \emph{reduces to evaluating a single CNN}. Furthermore, the pipeline is straightforward to implement on GPU, and is sufficiently memory-efficient to process multiple images at once. In practice, this results in an extremely fast detector which still retains excellent performance.

\paragraph{Dropping the SVM.} As discussed in Section~\ref{s:rcnn}, R-CNN involves training an SVM classifier for each target object class as well as fine-tuning the CNN features for all classes. An obvious question is whether SVM training is redundant and can be eliminated.

Recall from Section~\ref{s:rcnn} that fine-tuning learns a softmax predictor $\phi_{\text{sftmx}}$ on top of R-CNN features $\phi_{\rcnn}(\bx;R)$, whereas SVM  training learns a linear predictor $\phi_{\text{SVM}}$ on top of the same features. In the first case, $P_c = P(c | \bx, R) = [\phi_{\text{sftmx}} \circ \phi_{\rcnn}(\bx;R)]_c $ is an estimate of the class posterior for region $R$; in the second case $S_c = [\phi_{\text{SVM}}\circ \phi_{\rcnn}(\bx;R)]_c$ is a score that discriminates class $c$ from any other class (in both cases background is treated as one of the classes).  As verified in Section~\ref{s:experiments} and Table~\ref{tab:svm_vs_fc}, $P_c$ works poorly as a score for an object detector; however, and somewhat surprisingly, using as score the ratio $S'_c = P_c  / P_0$ results in performance nearly as good as using an SVM. Further, note that $\phi_{\text{sftmx}}$ can be decomposed as $C+1$ linear predictors $\langle w_c, \phi_{\rcnn}\rangle + b_c$ followed by exponentiation and normalisation; hence the scores $S'_c$ reduces to the expression $S'_c = \exp \left(\langle w_c - w_0, \phi_{\rcnn} \rangle + b_c - b_0\right)$.

\paragraph{Integrating SPP and bounding box regression.} While in the original implementation of SPP~\cite{he14spatial} the pooling mechanism is external to the CNN software, we implement it  directly as a layer $\SPP(\cdot;R_1,\dots,R_n)$. This layer takes as input a tensor representing the convolutional features $\cnv(\bx) \in \mathbb{R}^{H \times W \times D}$ and outputs $n$ feature fields of size $h \times w \times D$, one for each region $R_1,\dots,R_n$ passed as input. These fields can be stacked in a 4D output tensor, which is supported by all common CNN software. Given a dual CPU/GPU implementation of the layer, SPP integrates seamlessly with most CNN packages, with substantial benefit in speed and flexibility, including the possibility of training with back-propagation through it.

 Similar to SPP, bounding box regression is easily integrated as a bank of filters $(Q_c,b_c), c=1,\dots,C$ running on top of the convolutional features $\cnv(\bx)$. This is cheap enough that can be done in parallel for all the object classes.

\paragraph{Scale-augmented training, single scale evaluation.} While SPP is fast, one of the most time consuming step is to evaluate features at multiple scales~\cite{he14spatial}. However, the authors of~\cite{he14spatial} also indicate that restricting evaluation to a single scale has a marginal effect in performance. Here, we maintain the idea of evaluating the detector at test time by processing each image at a single scale. However, this requires the CNN to \emph{explicitly learn} scale invariance, which is achieved by fine-tuning the CNN using randomly rescaled versions of the training data.

\section{Experiments}\label{s:experiments}


\begin{table}[t]
	\centering
	\footnotesize
\begin{tabular}{| l | c c | c c | } \hline 
	Evaluation method & \multicolumn{2}{c|}{Single scale} & \multicolumn{2}{c|}{Multi scale} \\
	BB regression & no & yes & no & yes \\
	\hline
	$S_c$ (SVM)  & 54.0  & 58.6  & 56.3  & 59.7 \\ 
	$P_c$ (softmax) & 27.9  & 34.5  & 30.1  & 38.1 \\ 
	$P_c/P_0$ (modified softmax) & 54.0  & 58.0  & 55.3  & 58.4 \\ 
	\hline 
\end{tabular}
\vspace{1em}
\caption{Evaluation of SPP-CNN with and without the SVM classifier. The table report mAP on the PASCAL VOC 2007 test set for the single scale and multi scale detector, with or without bounding box regression. Different rows compare different bounding box scoring mechanism of Section~\ref{s:streamlined}: the SVM scores $S_c$, the softmax posterior probability scores $P_c$, and the modified softmax scores $P_c / P_0$.}
\label{tab:svm_vs_fc}
\end{table}

\begin{table*}[t]
	\centering
	\scriptsize
	\setlength{\tabcolsep}{1.5pt}
	\begin{tabular}[H]{| c | c | cccccccccccccccccccc | } \hline 
		method  & {\tiny mAP}   & {\tiny aero}   & {\tiny bike}   & {\tiny bird}   & {\tiny boat}   & {\tiny bottle}   & {\tiny bus}   & {\tiny car}   & {\tiny cat}   & {\tiny chair}   & {\tiny cow}   & {\tiny table}   & {\tiny dog}   & {\tiny horse}   & {\tiny mbike}   & {\tiny person}   & {\tiny plant}   & {\tiny sheep}   & {\tiny sofa}   & {\tiny train}   & {\tiny tv}  \\ \hline 
		SVM MS & \textbf{59.68}  & 66.8   & 75.8   & 55.5   & 43.1   & 38.1   & 66.6   & 73.8   & 70.9   & 29.2   & 71.4   & 58.6   & 65.5   & 76.2   & 73.6   & 57.4   & 29.9   & 60.1   & 48.4   & 66.0   & 66.8  \\ 
		SVM SS & \textbf{58.60}  & 66.1   & 76.0   & 54.9   & 38.6   & 32.4   & 66.3   & 72.8   & 69.3   & 30.2   & 67.7   & 63.7   & 66.2   & 72.5   & 71.2   & 56.4   & 27.3   & 59.5   & 50.4   & 65.3   & 65.2  \\ 
		\hline
		FC8 MS & \textbf{58.38}  & 69.2   & 75.2   & 53.7   & 40.0   & 33.0   & 67.2   & 71.3   & 71.6   & 26.9   & 69.6   & 60.3   & 64.5   & 74.0   & 73.4   & 55.6   & 25.3   & 60.4   & 47.0   & 64.9   & 64.4  \\ 
		FC8 SS & \textbf{57.99}  & 67.0   & 75.0   & 53.3   & 37.7   & 28.3   & 69.2   & 71.1   & 69.7   & 29.7   & 69.1   & 62.9   & 64.0   & 72.7   & 71.0   & 56.1   & 25.6   & 57.7   & 50.7   & 66.5   & 62.3  \\ 
		\hline
		FC8 C3k MS & \textbf{53.41}  & 55.8   & 73.1   & 47.5   & 36.5   & 17.8   & 69.1   & 55.2   & 73.1   & 24.4   & 49.3   & 63.9   & 67.8   & 76.8   & 71.1   & 48.7   & 27.6   & 42.6   & 43.4   & 70.1   & 54.5  \\ 
		FC8 C3k SS & \textbf{53.52}  & 55.8   & 73.3   & 47.3   & 37.3   & 17.6   & 69.3   & 55.3   & 73.2   & 24.0   & 49.0   & 63.3   & 68.2   & 76.5   & 71.3   & 48.2   & 27.1   & 43.8   & 45.1   & 70.2   & 54.6  \\ 
		\hline
	\end{tabular}
	\vspace{1em}
	\caption{Comparison of different variants of the SPP-CNN detector. First group of rows: original SPP-CNN using Multi Scale (MS) or Single Scale (SS) detection. Second group: the same experiment, but dropping the SVM and using the modified softmax scores of Section~\ref{s:streamlined}. Third group: SPP-CNN \emph{without region proposal generation}, but using a fixed set of 3K candidate bounding boxes as explained in Section~\ref{s:minusr}.}
	\label{tab:boxclusters}
\end{table*}

This section evaluates the changes to R-CNN and SPP-CNN proposed in Section~\ref{s:method}. All experiments use the Zeiler and Fergus (ZF) small CNN~\cite{zeiler14visualizing} as this is the same network used by~\cite{he14spatial} that introduce SPP-CNN. While more recent networks such as the very deep models of Simonyan and Zisserman~\cite{simonyan14very} are likely to perform better, this choice allows to compare directly~\cite{he14spatial}. The detector itself is trained and evaluated on the PASCAL VOC 2007 data~\cite{everingham07pascal}, as this is a default benchmark for object detection and is used in~\cite{he14spatial} as well.

\paragraph{Dropping the SVM.} The first experiment evaluates the performance of the SPP-CNN detector with or without the linear SVM classifier, comparing the bounding box scores $S_c$ (SVM), $P_c$ (softmax), and $S'_c$ (modified softmax) of Section~\ref{s:streamlined}. As can be seen in Table~\ref{tab:svm_vs_fc} and Table~\ref{tab:boxclusters}, the best performing method is SSP-CNN evaluated at multiple scales, resulting in 59.7\% mAP on the PASCAL VOC 2007 test data (this number matches the one reported in~\cite{he14spatial}, validating our implementation). Removing the SVM and using the CNN softmax scores directly performs really poorly, with a drop of 21.6\% mAP point. However, adjusting the softmax scores using the simple formula $P_c / P_0$ restores the performance almost entirely, back to 58.4\% mAP.  While there is still a small 1.3\% drop in mAP accuracy compared to using the SVM, removing the latter dramatically simplifies the detector pipeline, resulting in particular in significantly faster training as it removes the need of preparing and caching data for the SVM (as well as learning it).

\paragraph{Multi-scale evaluation.} The second set of experiments assess the importance of performing multi-scale evaluation of the detector. Results are reported once more in Tables~\ref{tab:svm_vs_fc} and~\ref{tab:boxclusters}. Once more, multi-scale detection is the best performing method, with performance up to 59.7\% mAP. However, single scale testing is very close to this level of performance, at 58.6\%, with a drop of just 1.1\% mAP points. Just like when removing the SVM, the resulting simplification and in this case detection speedup make this drop in accuracy more than tolerable. In particular, testing at a single scale accelerates detection roughly five-folds.

\begin{figure}
\vspace{1em}
	\centering
	\setlength{\figureheight}{2.3cm}
	\setlength{\figurewidth}{0.5\linewidth}
%
%
\definecolor{mycolor1}{rgb}{0.10588,0.61961,0.46667}%
\definecolor{mycolor2}{rgb}{0.85098,0.37255,0.00784}%
\definecolor{mycolor3}{rgb}{0.45882,0.43922,0.70196}%
\begin{tikzpicture}

\begin{axis}[%
width=\figurewidth,
height=\figureheight,
scale only axis,
xmin=0.5,
xmax=7.5,
xtick={1, 2, 3, 4, 5, 6, 7},
xlabel={Num Boxes/im [$10^3$]},
xmajorgrids,
ymin=0.35,
ymax=0.65,
ytick={0.3, 0.4, 0.5, 0.6},
ylabel={mAP},
ymajorgrids,
legend style={at={(1.03,0.5)},anchor=west,draw=black,fill=white,legend cell align=left},
minor tick num=1
]
\addplot [color=mycolor1,line width=1.0pt,mark size=1.5pt,only marks,mark=x,mark options={solid}]
  table[row sep=crcr]{%
2.004	0.56271527398063\\
};
\addlegendentry{SS};

\addplot [color=mycolor1,line width=1.0pt,mark size=1.5pt,only marks,mark=o,mark options={solid}]
  table[row sep=crcr]{%
2.004	0.596846047937723\\
};
\addlegendentry{SS-BBR};

\addplot [color=mycolor2,dashed,line width=1.0pt,mark size=1.5pt,mark=x,mark options={solid}]
  table[row sep=crcr]{%
1	0.437897031581719\\
2	0.45529901687878\\
3	0.45030128458836\\
5	0.458905846416616\\
7	0.460686783533161\\
};
\addlegendentry{Cx};

\addplot [color=mycolor2,solid,line width=1.0pt,mark size=1.5pt,mark=o,mark options={solid}]
  table[row sep=crcr]{%
1	0.507424634729652\\
2	0.529931590403993\\
3	0.534126281163669\\
5	0.543880415676741\\
7	0.544143682457749\\
};
\addlegendentry{Cx-BBR};

\addplot [color=mycolor3,line width=1.0pt,mark size=1.5pt,only marks,mark=x,mark options={solid}]
  table[row sep=crcr]{%
7	0.450645420150908\\
};
\addlegendentry{SW};

\addplot [color=mycolor3,line width=1.0pt,mark size=1.5pt,only marks,mark=o,mark options={solid}]
  table[row sep=crcr]{%
7	0.535700941310469\\
};
\addlegendentry{SW-BBR};

\end{axis}
\end{tikzpicture}%
	\caption{mAP on the PASCAL VOC 2007 test data as a function of the number of candidate boxes per image, proposal generation method, and using or not bounding box regression.}
	\label{fig:map_per_boxes}
\end{figure}
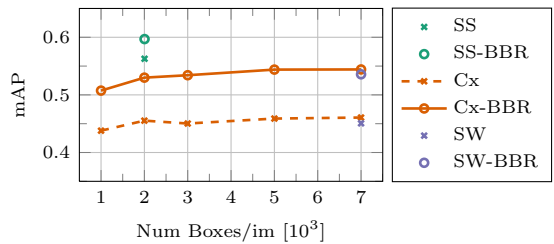

\begin{table*}[t]
		\scriptsize
		\centering
			\setlength{\tabcolsep}{1.5pt}
\begin{tabular}[H]{| l c | c | r r r r r r | r l | } \hline 
	\multicolumn{2}{|c|}{Impl. [ms]}  & SelS   & Prep.   & Move   & Conv   & SPP   & FC   & BBR   & \multicolumn{2}{|c|}{$\Sigma-\textrm{SelS}$}  \\ \hline 
	SPP & \multirow{2}{*}{MS} & \multirow{4}{*}{$1.98 \cdot 10^3$}  & $23.3$  & $67.5$ & $186.6$ & $211.1$ & $91.0$ & $39.8$ & $\mathbf{619.2}$ &  $\pm118.0 $ \\ 
	OURS& &  & $23.7$  & $17.7$ & $179.4$ & $38.9$ & $87.9$ & $9.8$ & $\mathbf{357.4}$ &  $\pm34.3 $ \\ \cline{1-2} \cline{4-11}
	SPP & \multirow{2}{*}{SS} &  & $9.0$  & $47.7$ & $31.1$ & $207.1$ & $90.4$ & $39.9$ & $\mathbf{425.1}$ &  $\pm117.0 $ \\ 
	OURS & &  & $9.0$  & $3.0$ & $30.3$ & $19.4$ & $88.0$ & $9.8$ & $\mathbf{159.5}$ &  $\pm31.5 $ \\ 
	\hline \end{tabular}
\caption{Timing (in $ms$) of the original SPP-CNN and our streamlined full-GPU implementation, broken down into selective search (SS) and preprocessing: image loading and scaling (Prep), CPU/GPU data transfer (Move), convolution layers (Conv), spatial pyramid pooling (SPP), fully connected layers (FC), and bounding box regression (BBR).}
\label{tab:timing}
\end{table*}

\paragraph{Dropping region proposal generation.} The next experiment evaluates replacing the SS region proposals $\mathcal{R}_{\text{SS}}(\bx)$ with the fixed proposals $\mathcal{R}_0(n)$ as suggested in Section~\ref{s:minusr}. Table~\ref{tab:boxclusters} shows the detection performance for $n=3,000$, a number of candidates comparable with the 2,000 extracted by selective search. While there is a drop in performance compared to using SS, this is small (59.68\% vs 53.41\%, i.e. a 6.1\% reduction), which is surprising since bounding box proposals are now oblivious of the image content.

Figure~\ref{fig:map_per_boxes} looks at these results in greater detail. Three bounding box generation methods are compared: selective search, sliding windows, and clustering (see also Section~\ref{s:minusr}), with or without bounding box regression. Neither clustering  nor sliding windows result in an accurate detector: even if the number of candidate boxes is increased substantially (up to $n=7K$), performance saturates at around 46\% mAP. This is much poorer than the $\sim$56\% achieved by selective search. Bounding box regression improves selective search by about 3\% mAP, up to $\sim$59\%, but it has a much more significant effect on the other two methods, improving performance by about 10\% mAP. Note that clustering with 3K candidates performs as well as sliding window with 7K.

We can draw several interesting conclusions. First, for the same low number of candidate boxes, selective search is much better than any fixed proposal set; less expected is that performance does not increase even with $3\times$ more candidates, indicating that the CNN is \emph{unable to tell which bounding boxes wrap objects better} even when tight boxes are contained in the shortlist of proposals. This can be explained by the high degree of geometric invariance in the CNN. At the same time, the CNN-based bounding box regressor can make loose bounding boxes significantly tighter, which requires geometric information to be preserved by the CNN. This apparent contradiction can be explained by noting that bounding box classification is built on top of the FC layers of the CNN, whereas bounding box regression is built on the convolutional ones. Evidently, geometric information is removed in the FC layers, but is still contained in the convolutional layers (see also Figure~\ref{f:splash}).

\paragraph{Detection speed.} The last experiment (Table~\ref{tab:timing}) evaluates the detection speed of SPP-CNN (which is already orders of magnitude faster than R-CNN) and our streamlined implementation. Not counting SS proposal generation, the streamlined implementation is between $1.7\times$ (multi-scale) to $2.6\times$ (single-scale) faster than original SPP, with the most significant gain emerging from the integrated SPP and bounding box regression implementation on GPU and consequent reduction of data transfer cost between CPU and GPU. 

As suggested before, however,  the bottleneck is selective search. Compared to the slowest MS SPP-CNN implementation of~\cite{he14spatial}, using all the simplifications of Section~\ref{s:method}, including removing selective search, results in an overall detection speedup of more than $16\times$, from about 2.5s per image down to 160ms (this at a reduction of about 6\% mAP points).

\section{Conclusions}\label{s:conclusions}

Our most significant finding is that current CNNs do contain sufficient geometric information for accurate object detection, although in the convolutional rather than fully connected layers. This finding opens the possibility of building state-of-the-art object detectors that rely exclusively on CNNs, removing region proposal generation schemes such as selective search, and resulting in integrated, simpler, and faster detectors.

Our current implementation of a proposal-free detector is already much faster than SPP-CNN, and very close, but not quite as good, in term of mAP. However, we have only begun exploring the design possibilities and we believe that it is a matter of time before the gap closes entirely. In particular, our current scheme is likely to miss small objects in the image. These may be retained by alternative methods to search the object pose space, such as for example Hough voting on top of convolutional features, which would maintain the computational advantage and elegance of integrated and streamlined CNN detectors while allowing to thoroughly search the image for object occurrences. 

\bibliographystyle{ieee}
\bibliography{local,/Users/vedaldi/src/bibliography/bibliography}

\begin{thebibliography}{10}\itemsep=-1pt

\bibitem{alexe10what}
B.~Alexe, T.~Deselaers, and V.~Ferrari.
\newblock What is an object?
\newblock In {\em Proc. {CVPR}}, 2010.

\bibitem{arbelaez14multiscale}
P.~Arbel{\'a}ez, J.~Pont-Tuset, J.~T. Barron, F.~Marques, and J.~Malik.
\newblock Multiscale combinatorial grouping.
\newblock In {\em Proc. {CVPR}}, 2014.

\bibitem{carreira12cpmc}
J.~Carreira and C.~Sminchisescu.
\newblock Cpmc: Automatic object segmentation using constrained parametric
  min-cuts.
\newblock In {\em {PAMI}}, 2012.

\bibitem{dalal05histograms}
N.~Dalal and B.~Triggs.
\newblock Histograms of oriented gradients for human detection.
\newblock In {\em Proc. {CVPR}}, 2005.

\bibitem{deng09imagenet}
J.~Deng, W.~Dong, R.~Socher, L.-J. Li, K.~Li, and L.~Fei-Fei.
\newblock {ImageNet: A Large-Scale Hierarchical Image Database}.
\newblock In {\em Proc. {CVPR}}, 2009.

\bibitem{everingham07pascal}
M.~Everingham, A.~Zisserman, C.~Williams, and L.~V. Gool.
\newblock The {PASCAL} visual obiect classes challenge 2007 ({VOC2007})
  results.
\newblock Technical report, Pascal Challenge, 2007.

\bibitem{felzenszwalb08a-discriminatively}
P.~F. Felzenszwalb, D.~McAllester, and D.~Ramanan.
\newblock A discriminatively trained, multiscale, deformable part model.
\newblock In {\em Proc. {CVPR}}, 2008.

\bibitem{girshick15fast}
R.~Girshick.
\newblock Fast {RCNN}.
\newblock In {\em arXiv}, number arXiv:1504.08083, 2015.

\bibitem{girshick14rich}
R.~B. Girshick, J.~Donahue, T.~Darrell, and J.~Malik.
\newblock Rich feature hierarchies for accurate object detection and semantic
  segmentation.
\newblock In {\em Proc. {CVPR}}, 2014.

\bibitem{gong14multi-scale}
Y.~Gong, L.~Wang, R.~Guo, and S.~Lazebnik.
\newblock Multi-scale orderless pooling of deep convolutional activation
  features.
\newblock In {\em Proc. {ECCV}}, 2014.

\bibitem{he14spatial}
K.~He, X.~Zhang, S.~Ren, and J.~Sun.
\newblock Spatial pyramid pooling in deep convolutional networks for visual
  recognition.
\newblock In {\em Proc. {ECCV}}, 2014.

\bibitem{he04multiscale}
X.~He, R.~Zemel, and M.~C.-P. {n}\'{a}n.
\newblock Multiscale conditional random fields for image labeling.
\newblock In {\em Proc. {CVPR}}, 2004.

\bibitem{hosang15what}
J.~Hosang, R.~Beneson, P.~Doll\'ar, and B.~Schiele.
\newblock What makes for effective detection proposals?
\newblock {\em arXiv:1502.05082}, 2015.

\bibitem{krizhevsky12imagenet}
A.~Krizhevsky, I.~Sutskever, and G.~E. Hinton.
\newblock Imagenet classification with deep convolutional neural networks.
\newblock In {\em Proc. {NIPS}}, 2012.

\bibitem{lazebnik06beyond}
S.~Lazebnik, C.~Schmid, and J.~Ponce.
\newblock Beyond bag of features: Spatial pyramid matching for recognizing
  natural scene categories.
\newblock In {\em Proc. {CVPR}}, 2006.

\bibitem{simonyan14very}
K.~Simonyan and A.~Zisserman.
\newblock Very deep convolutional networks for large-scale image recognition.
\newblock {\em CoRR}, abs/1409.1556, 2014.

\bibitem{sivic05discovering}
J.~Sivic, B.~C. Russel, A.~A. Efros, A.~Zisserman, and W.~T. Freeman.
\newblock Discovering objects and their location in images.
\newblock In {\em Proc. {ICCV}}, 2005.

\bibitem{uijlings13selective}
J.~Uijlings, K.~van~de Sande, T.~Gevers, and A.~Smeulders.
\newblock Selective search for object recognition.
\newblock {\em {IJCV}}, 2013.

\bibitem{vedaldi09multiple}
A.~Vedaldi, V.~Gulshan, M.~Varma, and A.~Zisserman.
\newblock Multiple kernels for object detection.
\newblock In {\em Proc. {ICCV}}, 2009.

\bibitem{viola01rapid}
P.~Viola and M.~Jones.
\newblock Rapid object detection using a boosted cascade of simple features.
\newblock In {\em Proc. {CVPR}}, 2001.

\bibitem{zeiler14visualizing}
M.~D. Zeiler and R.~Fergus.
\newblock Visualizing and understanding convolutional networks.
\newblock In {\em Proc. {ECCV}}, 2014.

\bibitem{zhao14cracking}
Q.~Zhao and Z.~L. an~B.~Yin.
\newblock Cracking bing and beyond.
\newblock In {\em Proc. {BMVC}}, 2014.

\bibitem{zitnick14edge}
C.~Zitnick and P.~Doll\'ar.
\newblock Edge boxes: Locating object proposals from edges.
\newblock In {\em Proc. {ECCV}}, 2014.

\end{thebibliography}
\end{document}